\documentclass[runningheads]{llncs}
\usepackage[T1]{fontenc}
\usepackage{amsmath}
\usepackage{subcaption}
\usepackage{url}
\usepackage{booktabs} %For better tables

\usepackage{graphicx}
\usepackage{import}
\begin{document}

\title{Latent Pollution Model: The Hidden Carbon Footprint in 3D Image Synthesis}
\titlerunning{Latent Pollution Model: Carbon Footprint in 3D Synthesis}
\author{Marvin Seyfarth\inst{1,3}\orcidID{0009-0007-4213-0750} \and
Salman Ul Hassan Dar\inst{1,2,3}\orcidID{0000-0002-7603-4245} \and
Sandy Engelhardt\inst{1,2,3}\orcidID{0000-0001-8816-7654 }}
\authorrunning{M. Seyfarth et al.}
\institute{Department of Internal Medicine III, Heidelberg University Hospital, Germany \\ \and
    AI Health Innovation Cluster (AIH), Germany \and
    German Centre for Cardiovascular Research (DZHK), Partner site Heidelberg/Mannheim, Germany \\
    \email{Seyfarth@stud.uni-heidelberg.de}
}
\maketitle              % typeset the header of the contribution

\begin{abstract}
Contemporary developments in generative AI are rapidly transforming the field of medical AI. 
These developments have been predominantly driven by the availability of large datasets and high computing power, which have facilitated a significant increase in model capacity.
Despite their considerable potential, these models demand substantially high power, leading to high carbon dioxide $(CO_2)$ emissions.
Given the harm such models are causing to the environment, there has been little focus on the carbon footprints of such models.
This study analyzes carbon emissions from 2D and 3D latent diffusion models (LDMs) during training and data generation phases, revealing a surprising finding: the synthesis of large images contributes most significantly to these emissions.
We assess different scenarios including model sizes, image dimensions, distributed training, and data generation steps. 
Our findings reveal substantial carbon emissions from these models, with training 2D and 3D models comparable to driving a car for 10 km and 90 km, respectively. The process of data generation is even more significant, with $CO_2$ emissions equivalent to driving 160 km for 2D models and driving for up to 3345 km for 3D synthesis. Additionally, we found that the location of the experiment can increase carbon emissions by up to 94 times, and even the time of year can influence emissions by up to 50\%. These figures are alarming, considering they represent only a single training and data generation phase for each model.
Our results emphasize the urgent need for developing environmentally sustainable strategies in generative AI.

\end{abstract}
\section{Introduction}

Recent developments in generative AI are revolutionizing medical imaging research.
Many downstream tasks that previously required extensive feature engineering appear more manageable through zero or few-shot learning using deep learning-based models \cite{khader2023medicaldiffusiondenoisingdiffusion}.
These advancements have been primarily driven by the availability of large datasets and enhanced computing power, enabling an increase in model capacity.
Consequently, training these models necessitates the use of highly power-consuming machines for extended durations.
Moreover, training a model is just one part of a very extensive process; other phases such as hyperparameter optimization and data synthesis, can be even more power-intensive.
This high power consumption results in significant carbon dioxide emissions, which can harm our environment. Carbon dioxide $(CO_2)$ is a greenhouse gas that traps heat in the Earth's atmosphere, contributing to global warming and climate change \cite{NOAA2024}. Increased levels of $CO_2$ lead to a rise in global temperatures, resulting in melting ice caps, rising sea levels, and more frequent and severe weather events such as hurricanes, droughts and floods \cite{EC2024}. Furthermore, elevated $CO_2$ levels can disrupt ecosystems, harm biodiversity, and negatively impact human health by worsening breathing problems and causing more heat-related illnesses \cite{EC2024}. These environmental and health impacts underscore the urgency of adressing carbon emissions from all sources, including those from the training and deployment of generative models in medical imaging.
Despite the significant impact of these emissions on our environment, it is surprising that there has been no comprehensive assessment of carbon emissions associated with generative models in medical imaging.

In this study, we perform a comprehensive evaluation of carbon emissions in unconditional latent diffusion models (LDMs). LDMs are foundational to some of the state-of-the-art generative AI models \cite{rombach2022highresolution} and provide a good baseline for assessing the carbon emissions of such models.  
We assess carbon emissions from 2D- and 3D-LDMs during the training and data generation phases. 
Additionally, we investigate various factors such as model size, image dimensions, distributed training, and sampling strategies to understand their impact on carbon emissions. 
We continuously monitor the power consumption by CPUs and GPUs using the Python tool 'Carbontracker' \cite{anthony2020carbontracker}.\\ 

\section{Methodology}
\subsection{Study Design}
We conducted experiments on both 2D and 3D latent diffusion models. LDMs are diffusion-based generative models that learn data generation in a low-dimensional latent space of an autoencoder. 
This approach facilitates the synthesis of high-resolution images, which is crucial in medical imaging research \cite{rombach2022highresolution}.
We opted for LDMs because they form the building blocks of many state-of-the-art generative AI models. Many complex multimodal models utilize pre-trained unconditional models or are trained in a hybrid manner with them.

To investigate how different factors in LDM training and data generation impact carbon emissions, we performed experiments in various settings and calculated the corresponding power consumption. These settings include:\\
\textbf{Model Size:}
Three architecture sizes were evaluated: small, medium, and large, differing mainly in the number of channels, residual blocks, and attention layers. \\
\textbf{Image Dimensions:}
Four image sizes were considered. In 2D-LDMs image dimensions were set as $192^2$, $256^2$, $320^2$, and $512^2$, and in 3D-LDMs image dimensions were set as $64^3$, $128^3$, $192^3$, and $256^2 \times 192$.\\
\textbf{Distributed Training:}
Models were trained both in a distributed manner and on a single GPU. Comparisons were made between models trained on a single GPU and those trained on four GPUs.\\
\textbf{Region-Wise Emission:}
Carbon emissions were compared across different countries by utilizing the respective carbon emission conversion factors.\\
\textbf{Data Generation Steps:}
In LDMs, data generation is based on an iterative denoising process. The impact of the number of iterations on the data generation process was also assessed.
\subsection{Network, Training, and Sampling}
The code and architectures were adapted from Pinaya \textit{et al.} \cite{pinaya2022brain}, with several modifications to enhance performance. The Variational Autoencoder was replaced with a Vector Quantised Generative Adversarial Network (VQ-GAN) \cite{tudosiu2022morphologypreserving}. The diffusion model was trained to optimize mean absolute error, the variance noise schedule was adjusted, and efficiency improvements were implemented in the code. To obtain the power consumption predictions for 150k training iterations and 10k synthesized samples, which are common values in the literature \cite{khader2023medicaldiffusiondenoisingdiffusion}, the mean and standard deviation of the components' power usage were first calculated based on 6000 training iterations and 50 synthesized samples. We utilized a pre trained VQ-GAN, focusing our power consumption and carbon emission analysis specifically on the diffusion model itself.
All experiments were conducted on a machine comprising two AMD Milan EPYC 7513 processors and a NVIDIA A100-SXM4-40GB GPU.
\subsection{Dataset}
All experiments utilized 120 in-house computed tomography (CT) scans. 
The volumes had dimensions of 512 pixels in width and height, with depths ranging from 220 to 409 pixels. The images were center cropped accordingly to match the image dimensions listed above.

\subsection{Carbon Emission tracking}
Typically, the machines on which the experiments are performed themselves do not emit noticeable greenhouse gases. 
Instead, the source of energy powering these machines contributes to greenhouse emissions. 
Therefore, emissions associated with a model can be implicitly monitored by tracking the model's power usage.
In this study, we focused on $CO_2$ because it is the most important anthropogenic greenhouse gas due to its proportion in the atmosphere \cite{noaaclimatechangeco2,epaglobalghgoverview}, and it is released during energy production in most of the current energy sources.

Power usage of CPUs and GPUs was monitored using the Carbontracker tool \cite{anthony2020carbontracker}, which integrates seamlessly with Python using minimal code. During training, monitored values were averaged over 50 epochs and utilized for final power consumption predictions. Similarly, during data generation, monitored values were averaged over 50 samples. Emissions are reported with their mean values and standard deviations.

Additionally, to account for the total energy consumption of equipment required for powering, cooling, and maintaining the data center, the emission values were further adjusted using a power usage effectiveness (PUE) factor of 1.55, as recommended by Bouza \textit{et al.} \cite{Bouza2023}.  It is calculated by dividing the total energy consumed by the facility by the energy consumed by the IT equipment.
To convert power consumption into $CO_2$ emission equivalents, further adjustments were made by multiplying the average carbon intensity of European countries in 2022, as reported by the European Environment Agency \cite{eea2024}, and the monthly German carbon intensity for 2023 from Electricity Maps \cite{electricitymaps2024}. 
The total carbon emission equivalents $CO_2e$ of the model can then be calculated as:
\begin{equation}
\label{eqn:CO2e}
CO_2e = P_{\text{total}} \times t_{\text{avg}} \times N \times \text{PUE} \times \text{CI} =  E \, [\text{kWh}]  \times \text{CI}
\end{equation}
where $P_{\text{total}}$ is the total power usage of the components (in Watts), $t_{\text{avg}}$ is the average duration of one epoch (in hours), $N$ is the number of epochs or synthesized samples, $\text{PUE}$ is the Power Usage Effectiveness, and $\text{CI}$ is the carbon emission conversion factor (grams of $CO_2$ per kilowatt-hour). $E$ is the predicted total energy consumed by the machine's components.

\section{Results}
\subsection{Image Dimensions} \label{subsec:Image_size}
To assess the impact of input image size on carbon emissions, power consumption during both training and synthesis was monitored across various dimensions. 
\begin{figure}[h!]
    \centering
    \begin{subfigure}[b]{0.49\textwidth}
        \centering
        \includegraphics[width=\textwidth,scale=2]{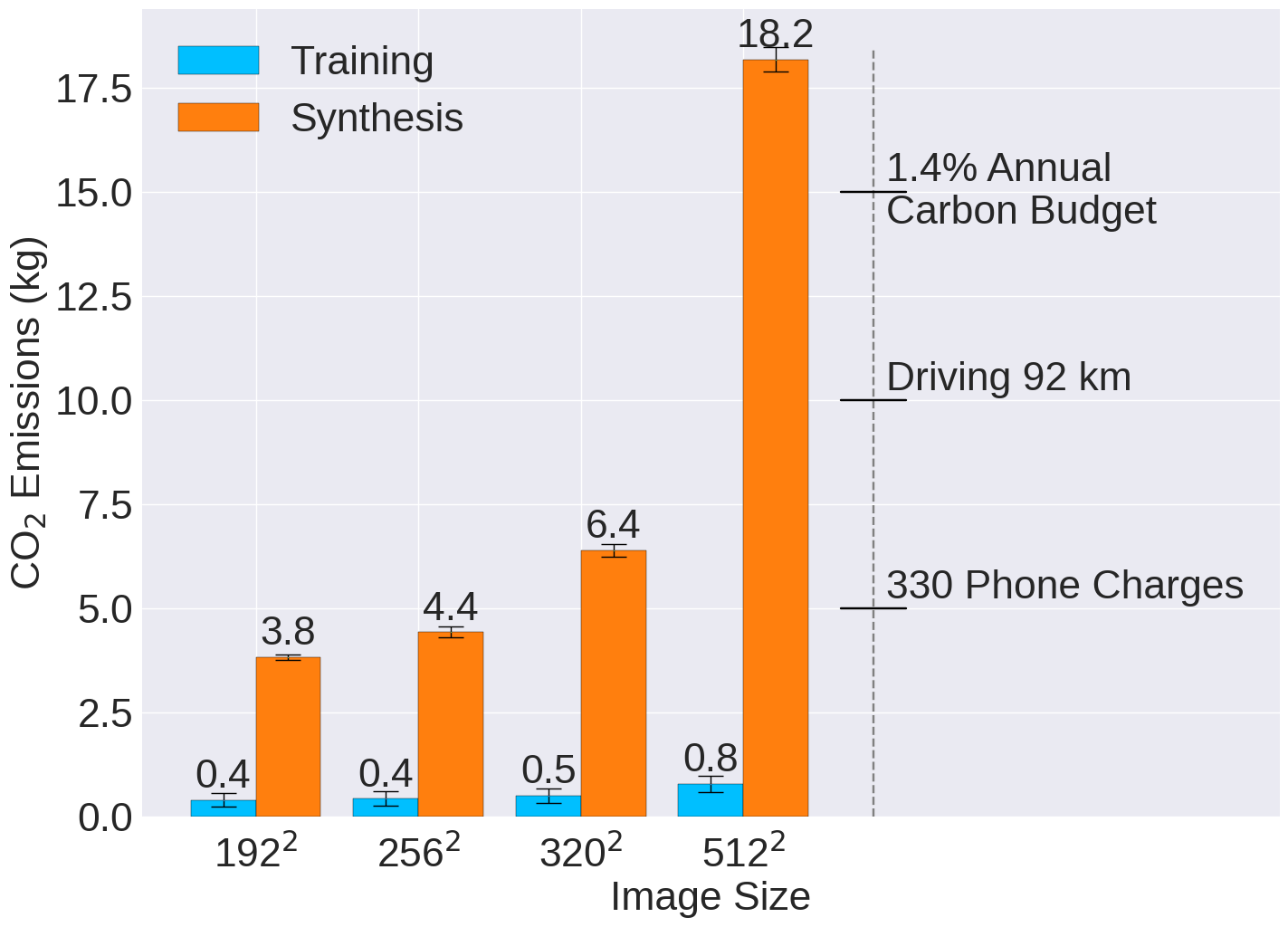}
        \caption{2D LDM}
        \label{fig:emission_2d_vol}
    \end{subfigure}
    \hfill
    \begin{subfigure}[b]{0.49\textwidth}
        \centering
        \includegraphics[width=\textwidth,scale=2]{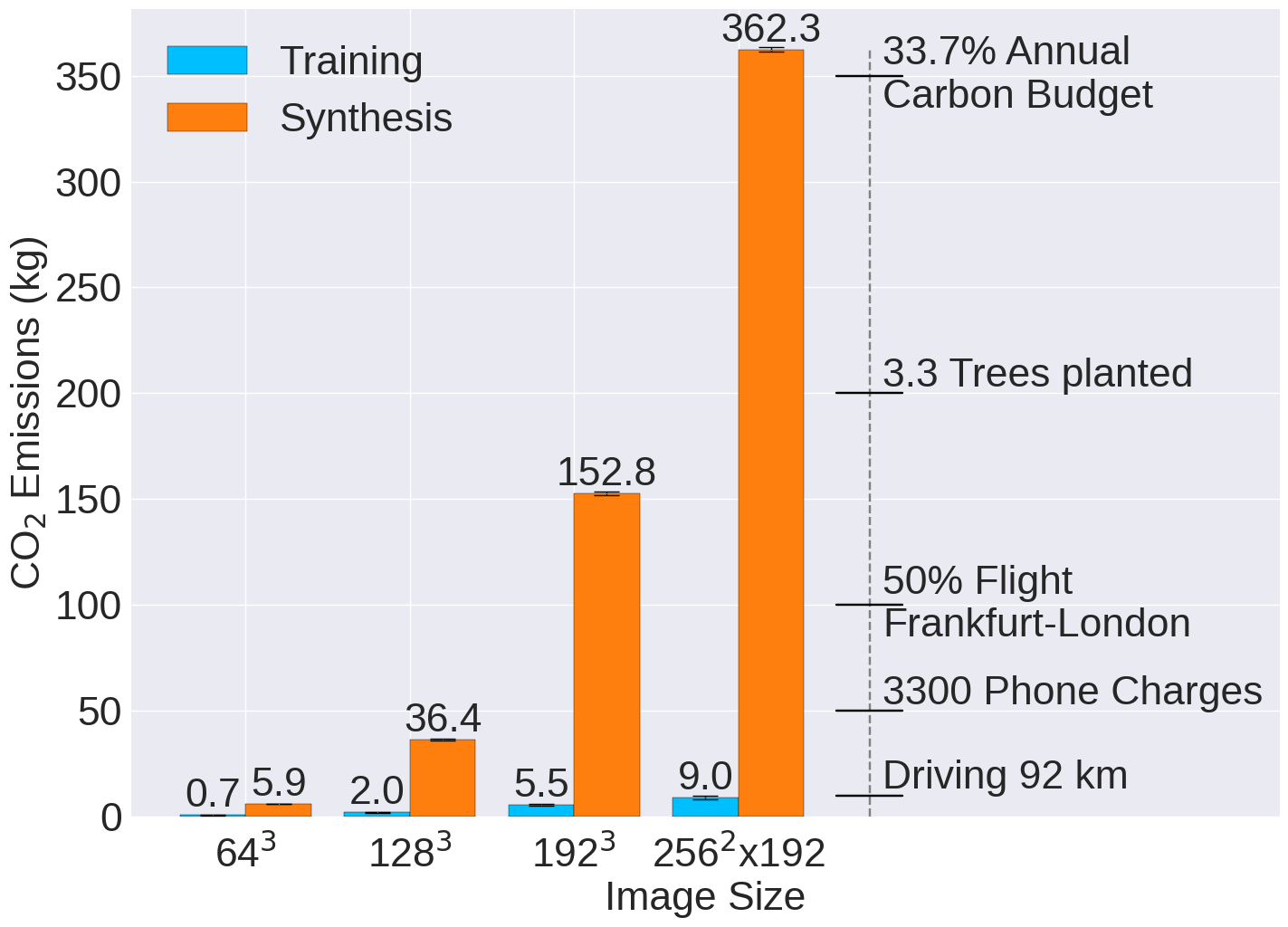}
        \caption{3D LDM}
        \label{fig:emission_3d_vol}
    \end{subfigure}
    \vfill
    \caption{$CO_2$ emissions from training and inference of 2D and 3D LDMs. The highest benchmark value represents the annual per capita carbon budget needed for a 50\% chance of limiting global warming to 1.5°C, as per the Paris Agreement \cite{unfcccparis2015}. For 3D synthesis, emissions reach 33.7\% of this budget, equivalent to a 1500km flight, such as from Tokyo to Seoul.}
    \label{fig:emissions_volume}
\end{figure}
Fig. \ref{fig:emissions_volume} shows the $CO_2$ emission equivalents calculated using Eq. \ref{eqn:CO2e} and the yearly average carbon intensity factor \( CI_{EU27}\), which represents all 27 EU countries, for each process. Emissions from training were estimated based on 150k iterations, while emissions from synthesis were calculated for 10k generated samples. A scale is provided to compare the $CO_2$ emissions with familiar activities and benchmarks. These comparisons include driving an average car in Europe (based on an emission factor of 108.2 g $CO_2$/km \cite{eea-emissions}), number of phone charges \cite{epa2023}, airplane distance traveled \cite{myclimate2023}, number of trees planted that grow for at least 10 years \cite{epa2023}, and the annual $CO_2$ budget per capita required to have a 50\% chance of staying below the 1.5°C global warming target \cite{lamboll2023}. \\
\\
\textbf{Training}: In 2D-LDMs, we observe that image size has minimal impact on carbon emissions, ranging from $(0.40 \pm 0.16)$ kg $CO_2e$ to $(0.79 \pm 0.20)$ kg $CO_2e$. There is no significant increase in emissions for the first three image sizes, with only a minimal increase observed at the largest image size.
In contrast, in 3D-LDMs, image size significantly impacts carbon emissions, with $CO_2$ emissions, ranging from $(0.74 \pm 0.22)$ kg $CO_2e $ to $(9.0 \pm 0.9)$ kg $CO_2e$ As resolution increases, so do the carbon emissions. For commonly selected 3D resolutions in medical imaging, the total carbon emissions are approximately an order of magnitude larger than those for 2D models. 
For instance, training a 3D latent diffusion model at a resolution of $256^2\times192$ pixels is equivalent to approximately 92 km of driving an average car in Europe, which is roughly equivalent to three days of driving for an average European citizen \cite{odyssee-mure}.\\
\textbf{Synthesis}: Although training leads to a considerable amount of carbon emissions, data synthesis emerges as a more power-intensive phase. This is due to the GPU operating at 100\% performance during synthesis, utilizing its full power capacity. The total carbon emissions are significantly determined by the shape of the synthesized image. In 2D-LDMs, carbon emissions range from $(3.83 \pm 0.07)$ kg $CO_2e$ to $(18.2 \pm 0.3)$ kg $CO_2e$ with energy usage increasing continuously as resolution increases.
The most substantial environmental impact is observed during the synthesis of medical 3D images, particularly at high resolutions. For 10k synthesized samples, the carbon emissions range from $(5.95 \pm 0.17)$ kg $CO_2e$ to $(362.3 \pm 1.0)$ kg $CO_2e$.

\subsection{Model Size}\label{subsec:Architecture_size}
\begin{figure}[t]
    \centering
    \begin{subfigure}[b]{0.49\textwidth}
        \centering
        \includegraphics[width=\textwidth,scale=2]{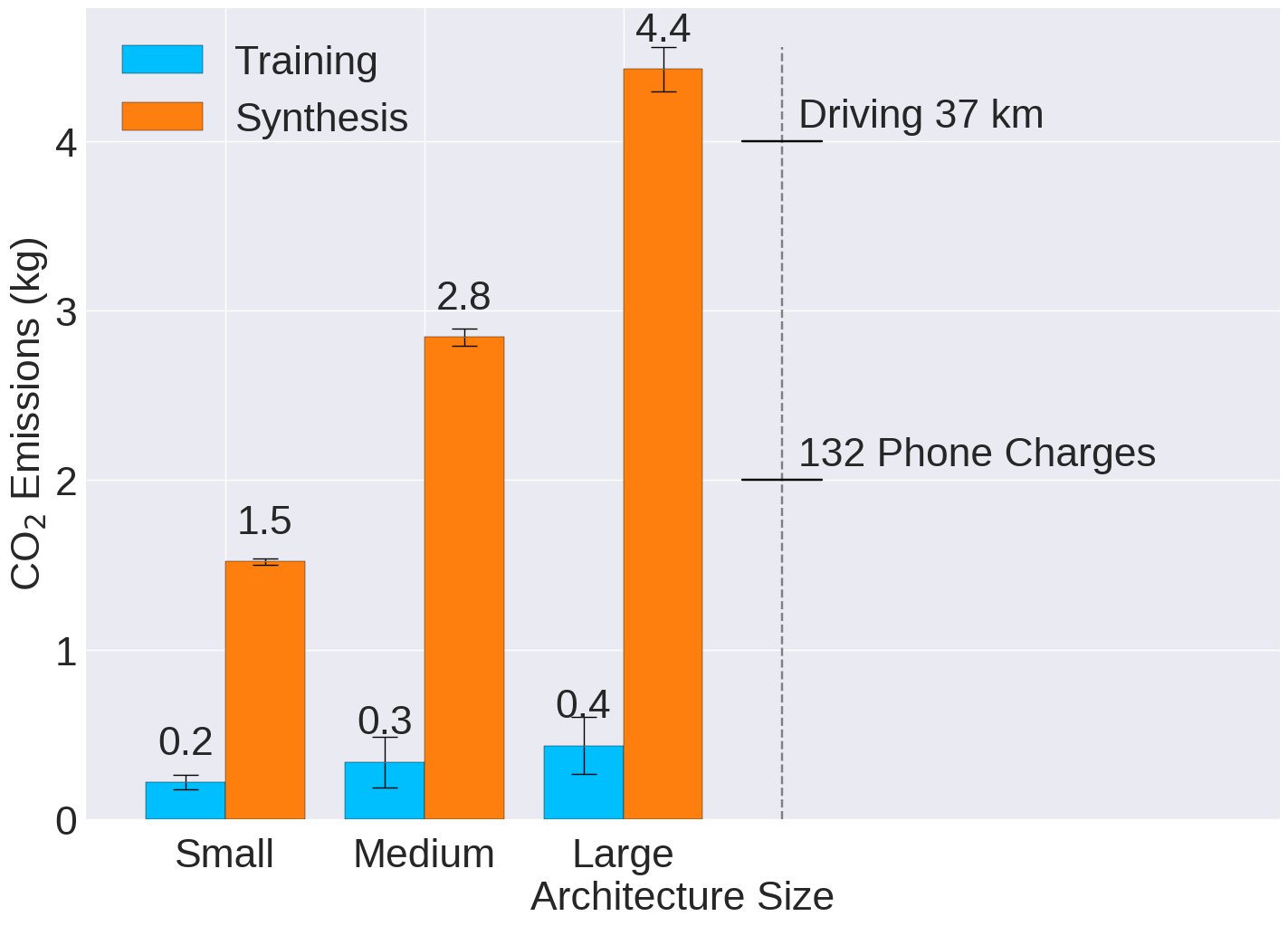}
        \caption{2D LDM}
        \label{fig:emission_2d_arch}
    \end{subfigure}
    \hfill
    \begin{subfigure}[b]{0.49\textwidth}
        \centering
        \includegraphics[width=\textwidth,scale=2]{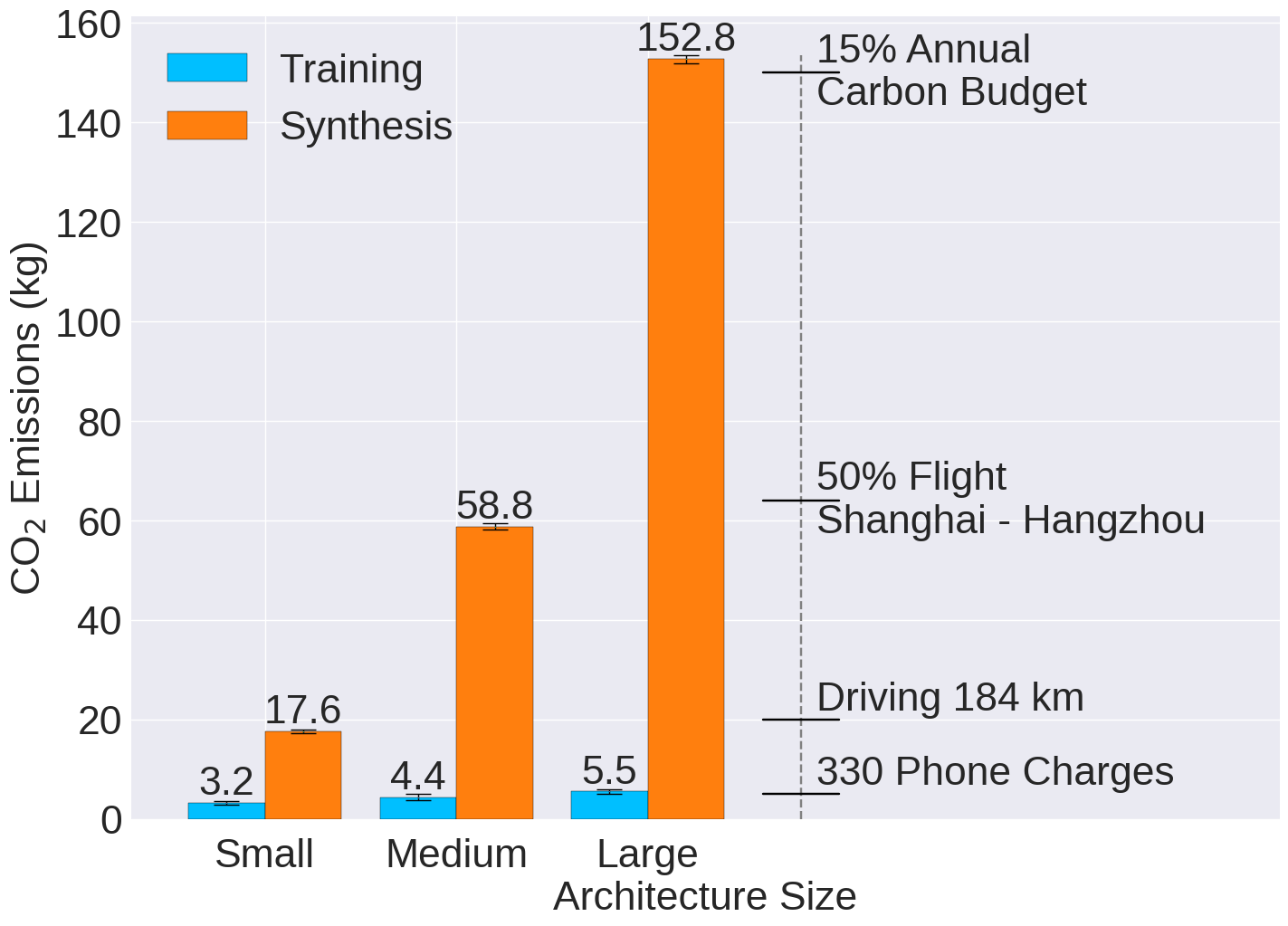}
        \caption{3D LDM}
        \label{fig:emission_3d_arch}
    \end{subfigure}
    \vfill
    \caption{$CO_2$ emissions by model architecture size in Training and Synthesis.}
    \label{fig:emission_architecture}
    \vspace{-10pt}
\end{figure}

As model depth and complexity increase, the number of trainable parameters rises, demanding more computational resources and time per iteration.  
In this study the image size of $256^2$ was used for 2D-LDMs and a volume size of $192^3$ was chosen for 3D models.
The carbon emissions as a function of model size, calculated using the European average emission factor \( CI_{EU27}\) are shown in Fig. \ref{fig:emission_architecture}.\\
\\
\textbf{Training}: As model complexity increases, the carbon emissions rise for both 2D- and 3D-models, with the model's size having a comparable influence in both cases. Specifically, as model complexity increases, carbon emissions rise by approximately 30\% to 40\%.\\
\textbf{Synthesis}: The influence of architecture size becomes more pronounced during synthesis, with carbon emissions increasing by 60\% to 80\% as model complexity grows. This highlights the significant role of model size not only in computational requirements but also in environmental impact during the synthesis phase.
\subsection{Regional and Temporal influence}\label{subsec:Individual_factors}
 
In the preceding sections, experiments utilized the yearly average carbon 
intensity factor \( CI_{EU27}\), representing all 27 EU countries. 
\begin{table}[h!]
\centering
\caption{Yearly average carbon intensity factors for various regions.}
\begin{tabular}{lcccccc}
\toprule
\textbf{Region} & \textbf{EU27} & \textbf{Sweden} & \textbf{Germany} & \textbf{Poland} & \textbf{China} & \textbf{USA} \\
\midrule
\textbf{CI [gCO$_2$/kWh]} & 251 \cite{eea2024} & 7 \cite{eea2024} & 366 \cite{eea2024} & 666 \cite{eea2024} & 582 \cite{statista2023} & 390 \cite{eia2023} \\
\bottomrule
\end{tabular}

\label{tab:carbon_intensity}
\end{table}
Additionally, carbon emissions were calculated using the 2022 yearly average CI factors for specific regions as listed in Tab. \ref{tab:carbon_intensity}.

\begin{figure}
    \centering
    
    \begin{subfigure}[b]{0.49\textwidth}
        \centering
        \includegraphics[width=\textwidth,scale=2]{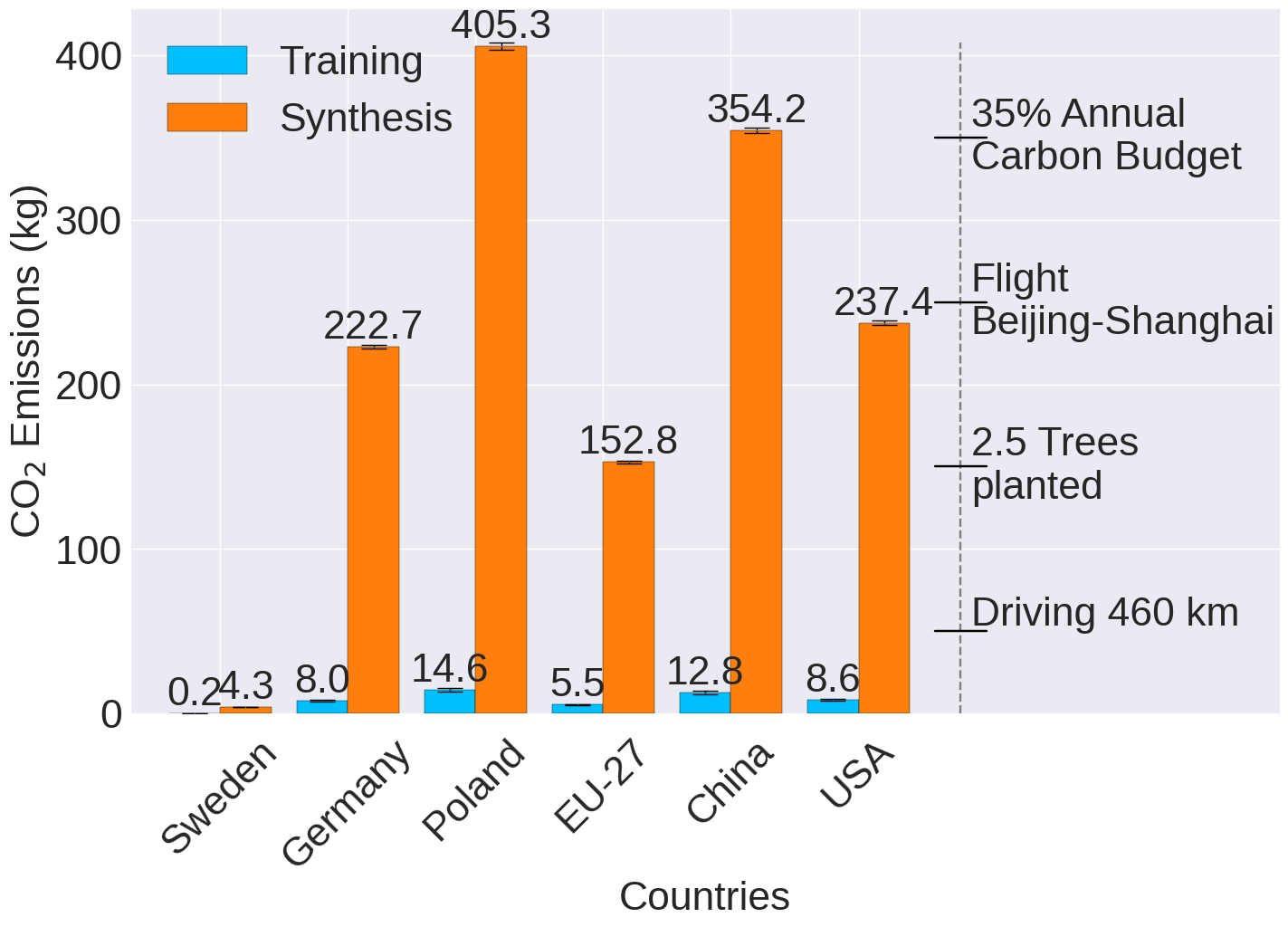}
        \caption{Regional Variations in CO\textsubscript{2} Emissions for Training and Synthesis.}
        \label{fig:regional_differences}
    \end{subfigure}
    \hfill
    \begin{subfigure}[b]{0.49\textwidth}
        \centering
        \includegraphics[width=\textwidth,scale=2]{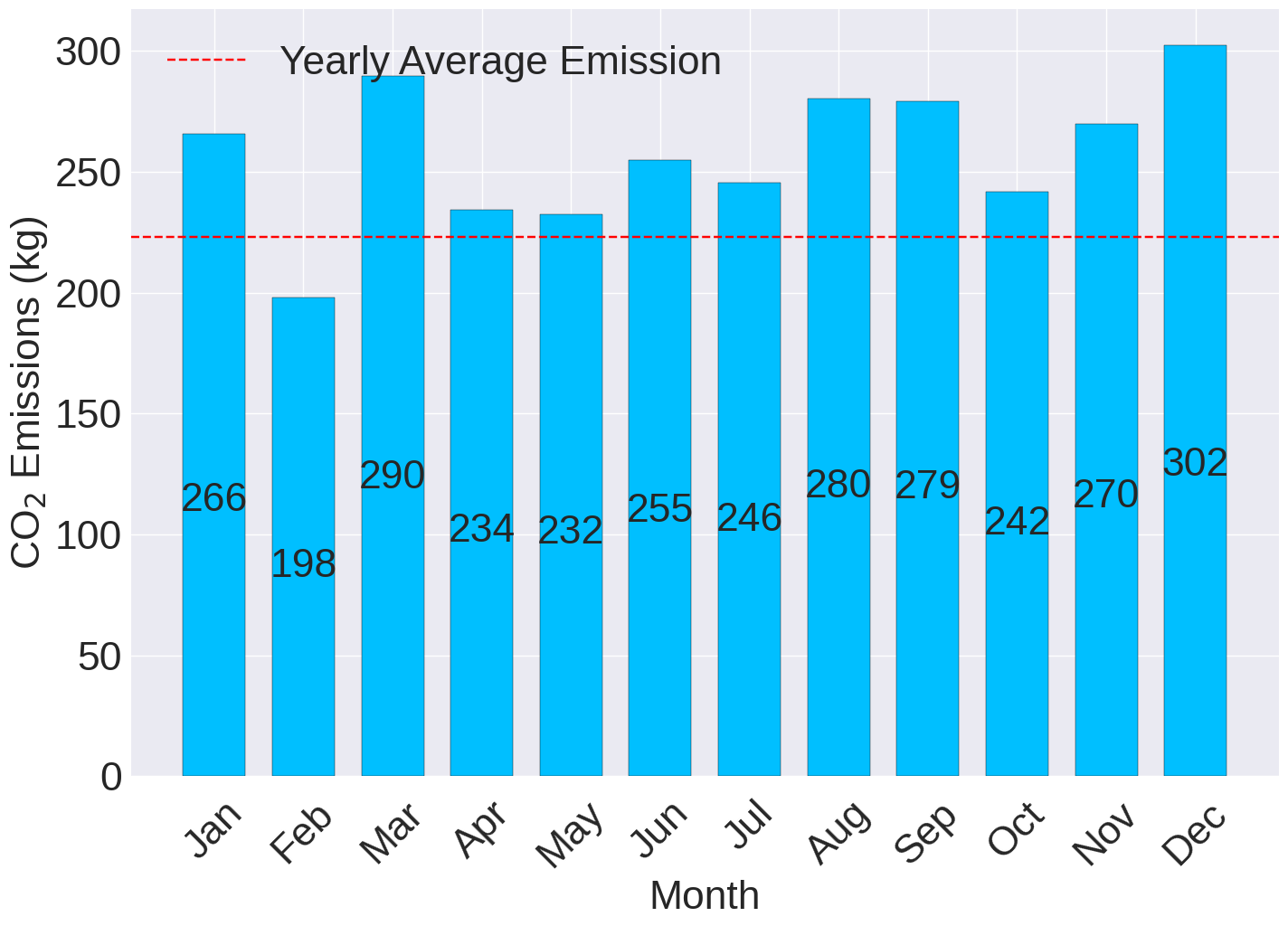}
        \caption{Monthly CO\textsubscript{2} emissions for Synthesis in 2022 (Germany).}
        \label{fig:temporal_differences}
    \end{subfigure}
    
    \caption{Regional and temporal differences in Carbon Emissions in a 3D LDM.}
    \vspace{-15pt}
    \label{fig:mainfig}
\end{figure}

Fig. \ref{fig:regional_differences} illustrates how the region where the experiment is conducted significantly influences total carbon emissions. For instance, the same experiments result in 4.3 kg $CO_2$e in Sweden and up to 405 kg $CO_2e$ in Poland, which exceeds 35\% of the annual $CO_2$ budget per capita to stay below 1.5°C of global warming.
Another influencing factor, although less significant, is the time of the year when experiments are conducted. Fig. \ref{fig:temporal_differences} shows the monthly distribution of carbon emissions for Germany in 2022. Variations of up to 50\% in carbon emissions can be observed between different months (e.g., December vs February).

\subsection{Reducing Model's Carbon Footprint}\label{subsec:Reduce_carbon}

\begin{figure}[h!]
    \centering
    \begin{subfigure}[b]{0.49\textwidth}
        \centering
        \includegraphics[width=\textwidth,scale=2]{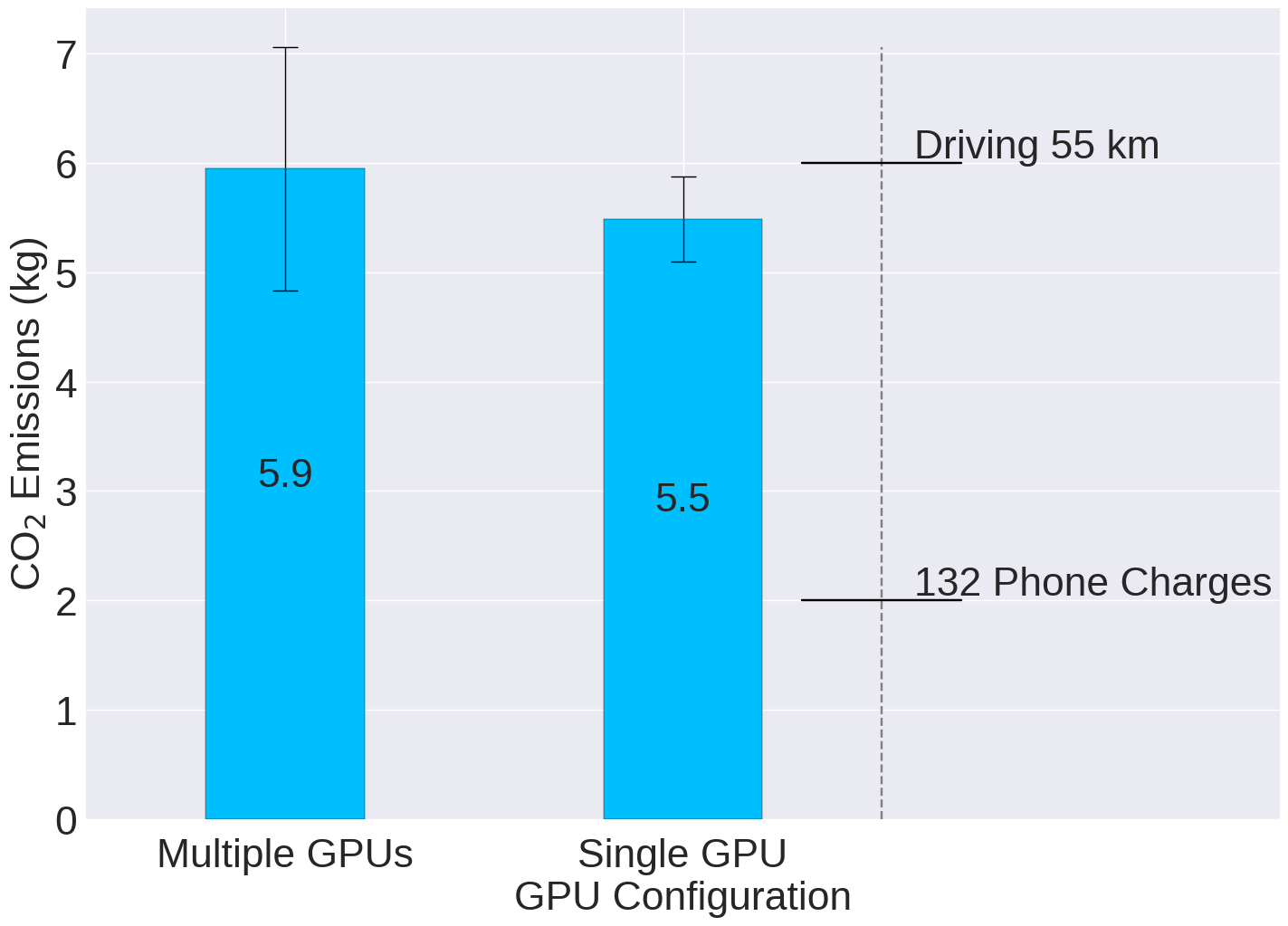}
        \caption{$CO_2$ emissions by GPU configuration in Training using \( CI_{EU27}\).}
        \label{fig:emission_ddp}
    \end{subfigure}
    \hfill
    \begin{subfigure}[b]{0.49\textwidth}
        \centering
        \includegraphics[width=\textwidth,scale=2]{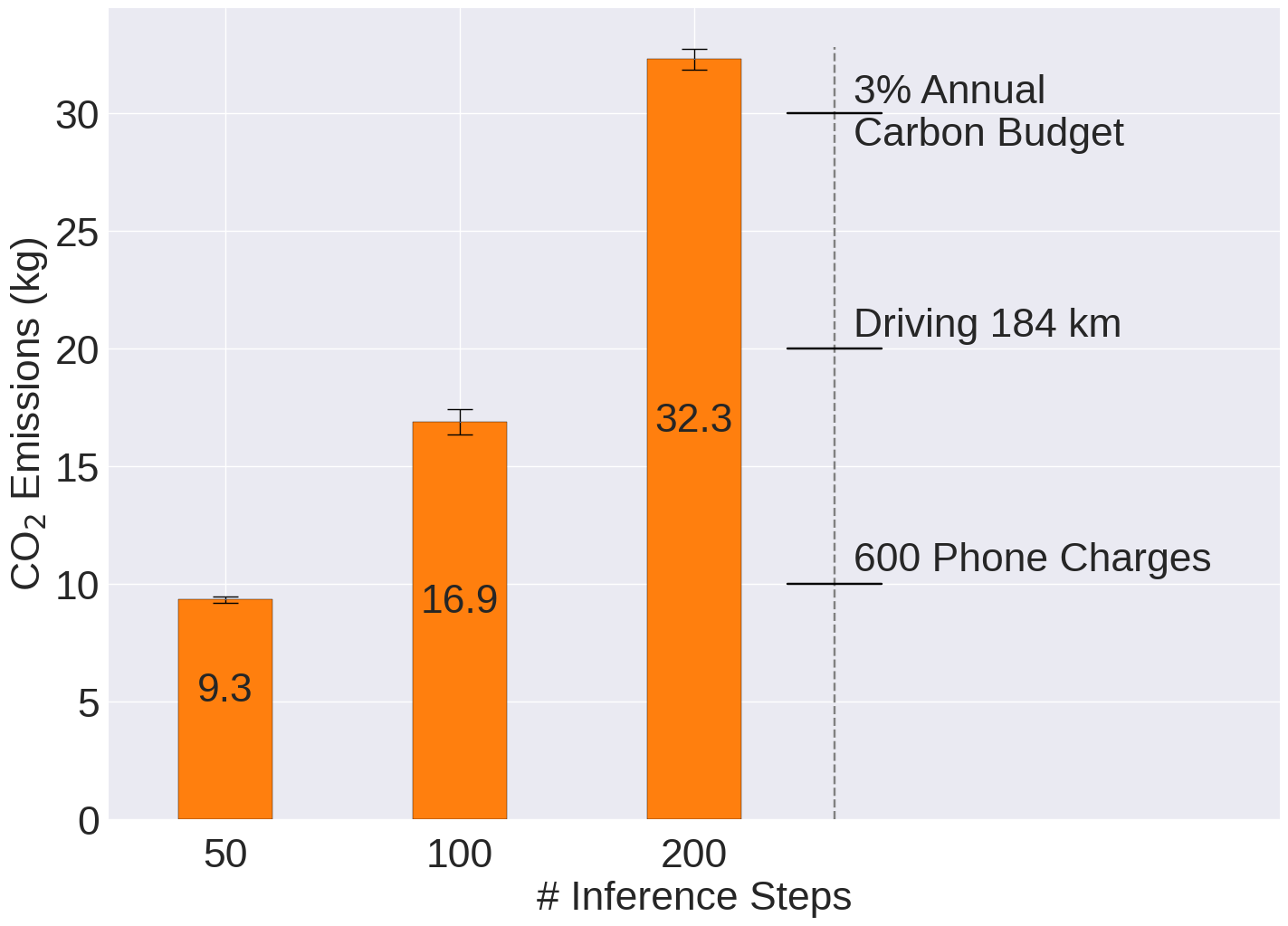}
        \caption{$CO_2$ emissions by denoising steps in Synthesis using \( CI_{EU27}\).}
        \label{fig:emission_steps}
    \end{subfigure}

    \caption{$CO_2$ emissions of different speed-up techniques in 3D LDM.}
    \label{fig:emission_speed_up}
    \vspace{-15pt}
\end{figure}

To evaluate strategies for reducing carbon emissions, we conducted experiments using distributed training and varying numbers of inference steps during synthesis in diffusion models. In this section the European average carbon intensity factor \( CI_{EU27}\) is used.\\
\textbf{Distributed Training:}
We compared emissions when training with 4 GPUs versus a single GPU, each processing the same batch size. Fig. \ref{fig:emission_ddp} illustrates the comparison.  Interestingly, for the 3D LDM, there was no significant reduction in carbon emissions observed with distributed training.\\
\textbf{Inference Steps in Synthesis:}
Fig. \ref{fig:emission_steps} shows the impact of varying the numbers of inference steps on carbon emissions during synthesis. It is evident that reducing the number of synthesis steps leads to a notable reduction in carbon emissions. For instance, opting for 200 inference steps instead of 1000 
can reduce emissions by 80\%, as shown in comparison with Fig. \ref{fig:emission_3d_vol}. This finding suggests that optimizing algorithms to reduce synthesis steps not only enhances synthesis efficiency but also significantly lowers the carbon footprint of the model.

\section{Discussion}
In this study, we investigated the carbon emissions associated with latent diffusion models by varying parameters during model training and data synthesis. Our findings reveal significant environmental implications, particularly in the context of 3D-LDMs and during the synthesis phase, where carbon emissions were notably high.\\
\textbf{Environmental Impact of LDMs:}
We observed alarmingly high carbon emissions, especially in 3D-LDMs and during data synthesis.
Synthesizing 10k samples alone can consume over 34\% of the annual $CO_2$ budget per capita required to stay below the 1.5°C global warming target. This highlights the substantial environmental impact of generating artificial 3D datasets.
In our study, the carbon emissions during training reflect the emissions from a single experiment consisting of 150k training iterations. However, in practical settings, additional experiments, such as hyperparameter tuning and further data synthesis to evaluate model performance, contribute significantly to overall emissions.  
The scale of data generation can be immense; some studies generate hundreds of thousands of synthetic images as publicly available datasets  \cite{pinaya2022brain}. Moreover, the prevalence of patient data copies in synthesized datasets \cite{dar2024unconditionallatentdiffusionmodels,dar24memorization} implies that the actual number of uniquely generated samples is substantially higher, exacerbating the environmental impact.\\
\textbf{Mitigation Strategies:}
Our analysis suggests that while conventional practices like using multiple GPUs for training do not significantly reduce carbon emissions due to increased power consumption, reducing the number of inference steps during synthesis, albeit potentially compromising image quality, may be a justified approach to mitigate emissions during inference. Furthermore, general considerations like the timing and location of the experiment can significantly reduce the researchers' carbon footprint, as demonstrated in this study.\\
\textbf{Awareness and Sutainability:}
Raising awareness about the carbon footprint of generative AI models and advocating for sustainable practices in research is crucial. Researchers often overlook the environmental impact of their experiments, focusing solely on achieving scientific goals. Especially in research centers, there are very few limitations in terms of power usage and users are not directly aware of the actual used resources.
Incorporating tools such as Carbontracker into workflows can provide insights into power consumption and carbon emissions, encouraging sustainable model development and deployment.

Given the substantial carbon emissions generated during both training and data generation phases, our study underscores the importance of integrating environmental considertations into medical AI research practices. Promoting awareness and implementing tools for carbon emission tracking can contribute to more sustainable advancements in AI technology.

\subsubsection{Acknowledgment:} This work was supported through state funds approved by the State Parliament of Baden-Württemberg for the Innovation Campus Health + Life Science Alliance Heidelberg Mannheim, BMBF-SWAG Project 01KD2215D, and Informatics for life project through Klaus Tschira Foundation and by Multi-DimensionAI project of the Carl Zeiss Foundation. The authors also gratefully acknowledge the data storage service SDS@hd supported by the Ministry of Science, Research and the Arts  Baden-Württemberg (MWK) and the German Research Foundation (DFG) through grant INST 35/1314-1 FUGG and INST 35/1503-1 FUGG. The authors also acknowledge support by the state of Baden-Württemberg through bwHPC
and the German Research Foundation (DFG) through grant INST 35/1597-1 FUGG. We would like to thank O. Nerjes and M. Baumann for their valuable insights on the hardware configuration of the Helix Cluster. We also extend our gratitude to R. Karl for his input on CO\textsubscript{2} emissions in generative AI. \\
%\newpage
\bibliographystyle{splncs04}
\bibliography{references}
\newpage

\end{document}